\documentclass{article}

\PassOptionsToPackage{numbers}{natbib}

\usepackage[final]{template/neurips_2024}
\usepackage[utf8]{inputenc} 
\usepackage[T1]{fontenc}    
\usepackage{hyperref}       
\usepackage{url}            
\usepackage{booktabs}       
\usepackage{amsfonts}       
\usepackage{nicefrac}       
\usepackage{microtype}      
\usepackage{xcolor}         
\usepackage{blkarray, bigstrut}
\usepackage{gauss}
\usepackage{graphicx}
\usepackage{multirow}
\usepackage{amsmath}
\usepackage{bbm}
\usepackage{graphicx}
\usepackage{subcaption}
\newcommand*{\affmark}[1][*]{\textsuperscript{#1}}

\title{Black-box Uncertainty Quantification Method for LLM-as-a-Judge}
\author{ { \bf Nico Wagner\affmark[1], Michael Desmond\affmark[1], Rahul Nair\affmark[1], Zahra Ashktorab\affmark[1],}\\
  {\bf Elizabeth M. Daly\affmark[1], Qian Pan\affmark[1], Martín Santillán Cooper\affmark[1], }\\
  {\bf James M. Johnson\affmark[1], Werner Geyer\affmark[1]}\\
  \affmark[1]IBM Research\\
  \texttt{\{nico.wagner@, mdesmond@us., rahul.nair@ie.,}\\
  \texttt{zahra.ashktorab1@, elizabeth.daly@ie., qian.pan@,}\\
  \texttt{msantillancooper@, jmjohnson@us., werner.geyer@us.\}ibm.com} \\}

\begin{document}
\maketitle

\begin{abstract}
  LLM-as-a-Judge is a widely used method for evaluating the performance of Large Language Models (LLMs) across various tasks. We address the challenge of quantifying the uncertainty of LLM-as-a-Judge evaluations. While uncertainty quantification has been well-studied in other domains, applying it effectively to LLMs poses unique challenges due to their complex decision-making capabilities and computational demands. In this paper, we introduce a novel method for quantifying uncertainty designed to enhance the trustworthiness of LLM-as-a-Judge evaluations. 
  The method quantifies uncertainty by analyzing the relationships between generated assessments and possible ratings. By cross-evaluating these relationships and constructing a confusion matrix based on token probabilities, the method derives labels of high or low uncertainty. We evaluate our method across multiple benchmarks, demonstrating a strong correlation between the accuracy of LLM evaluations and the derived uncertainty scores. Our findings suggest that this method can significantly improve the reliability and consistency of LLM-as-a-Judge evaluations.
\end{abstract}

\section{Introduction}

Large Language Models (LLMs) have become integral to a wide range of tasks, including question-answering \cite{singhal2023towards}, summarization \cite{jin2024comprehensive}, translation \cite{zhang2023prompting}, concept extraction \cite{fang2022data}, classification \cite{howard2018universal}, and reasoning \cite{huang2022towards}. The evaluation of the texts they generate has emerged as a significant challenge due to data contamination \cite{balloccu-etal-2024-leak}, replicability, and the use of standard metrics or benchmarks \cite{liang2022holistic, hendrycks2020measuring, ghazal2013bigbench} which may not cover all dimensions of use case-specific evaluations.

An emerging method for evaluating generated content involves using other LLMs as evaluators, a method referred to as LLM-as-a-Judge \cite{zheng2024judging}. These evaluations can take various forms, including explanations, numeric values, or categorical ratings. This work specifically focuses on LLM-as-a-Judge methods that employ categorical ratings or numerical evaluations to assess generated outputs.

Despite their widespread use, LLM-as-a-Judge methods do not always align with human judgments, leading to instances where the evaluations may be incorrect or misleading \cite{wang2023large,bavaresco2024llms}. This divergence highlights the need to assess the trustworthiness of LLM-generated evaluations. Various techniques have been proposed to enhance the performance of LLMs and improve the reliability of their judgments \cite{verga2024replacing}.

To improve trustworthiness in LLM-as-a-Judge evaluations and to leverage the strengths of these methods, we introduce a novel approach called \emph{confusion-based uncertainty}. Our method is designed to quantify the uncertainty associated with LLM evaluations where evaluation outcomes are discrete, i.e. multiple choice settings, or fixed number of output options. This encompasses a majority of typical evaluation tasks included those involving human evaluations. 

\begin{figure}
  \centering
  \includegraphics[scale=0.55]{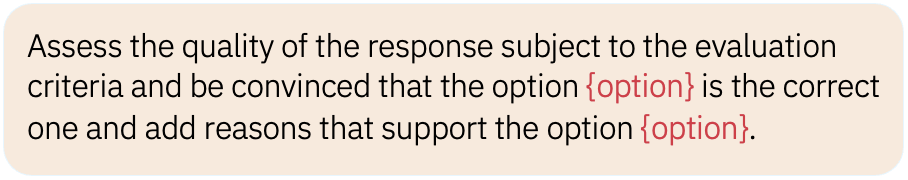}
  \caption{A biased assessment prompt. The LLM is prompted to asses a response (an input text that is under evaluation) under the assumption that a particular output option (label) is correct. By producing biased assessments, it is possible to determine the LLM's belief in a correct output option subject to assessments that may be contrary to this belief.}\label{fig:assessment_prompt}
\end{figure}

The \emph{confusion-based uncertainty} approach, inspired by chain-of-thought reasoning \cite{wei2022chain} and confusion matrices \cite{enwiki:1238399299}, first prompts the judge LLM to generate an assessment for each potential output option, biased on the implication that the option is correct. An assessment is an open ended evaluation produced by the LLM prior to making a final judgement. A biased assessment is generated under the implication that a given option is correct (see Figure \ref{fig:assessment_prompt}). For each biased assessment, the probability of all output options is recorded using log probabilities. This facilitates an analysis of the relationship between the biased assessments and the LLMs belief in a particular output being correct. A confusion matrix is constructed from these combinations, and levels of uncertainty are derived from the confusion matrix by looking at the distribution of token probabilities for output options, subject to each of the biased assessments. If an option is consistently likely across all potentially biased assessments, it is deemed to have low uncertainty.

The goal of \emph{confusion-based uncertainty} is to label LLM-as-a-Judge evaluations with high or low uncertainty, offering a clear signal of the evaluations' likely accuracy. 
We empirically evaluate our confusion-based uncertainty method across diverse benchmarks and models. Our results indicate that low uncertainty ratings correlate with higher accuracy, and the method effectively transfers across datasets and models.

\section{Related Work}

The capabilities of large language models have rapidly advanced, leading to their application in increasingly complex tasks \cite{srivastava2022beyond}. However, the growing sophistication of these models also raises new challenges, particularly in evaluating their outputs and understanding their uncertainty \cite{hu2023uncertainty}. As models are increasingly used as evaluators, what is often referred to as LLM-as-a-Judge, it becomes essential to develop methods that allow these models not only to generate responses but also to assess the confidence and reliability of those responses.

A promising line of research addresses these challenges through methods like chain-of-thought reasoning \citep{wei2022chain} and self-reflection \cite{ji2023towards}. Chain-of-thought reasoning enhances an LLM's ability to arrive at more accurate conclusions by breaking down complex tasks into intermediate logical steps. By guiding the model through a series of smaller, connected reasoning steps, the model’s decision-making process becomes more transparent and robust. Self-reflection, on the other hand, encourages the model to review and critique its own responses, iterating upon initial outputs to refine and improve its accuracy. Both techniques align with the broader framework of agentic design patterns \cite{NgAndrewAgenticDesignPatterns2024}, which foster active engagement by the model in evaluating and refining its generated outputs.

These reasoning-based approaches are particularly relevant to the growing body of work on LLM-as-a-Judge, where LLMs are used to evaluate data and provide judgments that are comparable to human annotations. Several works have studied LLM-as-a-Judge with evaluations generally focused on correlations between labels generated by LLMs and human annotations. For example, \citep{zheng2024judging, kim2023prometheus} and report strong agreements with human annotations, while other studies have reported mixed results \citep{doddapaneni2024finding, bavaresco2024llms}. Some works have proposed using ensembles of smaller models to increase performance \cite{verga2024replacing}, while others have used instruction fine-tuning to build customised evaluators \cite{kim2023prometheus}. 

Several works have explored methods for uncertainty estimation in the context of large language models. One approach is calibration-based uncertainty quantification, introduced by \cite{shen2024thermometer}, which focuses on efficient calibration using a auxiliary model trained over multiple tasks. However, this approach relies on internal model representations to produce features. In contrast, black-box uncertainty quantification methods, which do not require access to the internal workings of the model, have also emerged. 

Lin etal. \cite{lin2022teaching} investigated prompting strategies that guide LLMs to verbalize their uncertainty, especially when fine-tuned with labeled confidence values. Their work demonstrates how external methods can elicit uncertainty without accessing internal model states. 

Xiong et al. \cite{xiong2023can} evaluated several prompting strategies for eliciting uncertainty in LLMs, including direct assessment, chain-of-thought reasoning, and self-probing. Their findings indicate that models tend to exhibit overconfidence, particularly in general settings. To address this, they proposed strategies such as prompt perturbation, paraphrasing, and entity amplification, which reduced overconfidence and improved uncertainty predictions. They also developed a logistic regression model to predict uncertainty based on these perturbations \cite{pedapati2024large}. 

Kuhn et al. \cite{kuhn2023semantic} introduced semantic entropy as a novel approach to capture uncertainty by identifying semantically equivalent prompts. By measuring the variation in responses to these semantically similar prompts, their method provides a more nuanced understanding of model uncertainty.

\section{Confusion-based Uncertainty}
\label{sec:method}

In many LLM-as-a-Judge frameworks, the evaluation of generated text is conducted against predefined criteria \cite{kim2024prometheus}. Each criterion consists of a question and a set of options, among which the LLM must choose. The questions can vary widely, and the options can be defined as numeric values, words, or any other format, with no restriction on the number of words or the nature of the options.

\begin{figure}[ht]
  \centering
  \fbox{\includegraphics[scale=0.33]{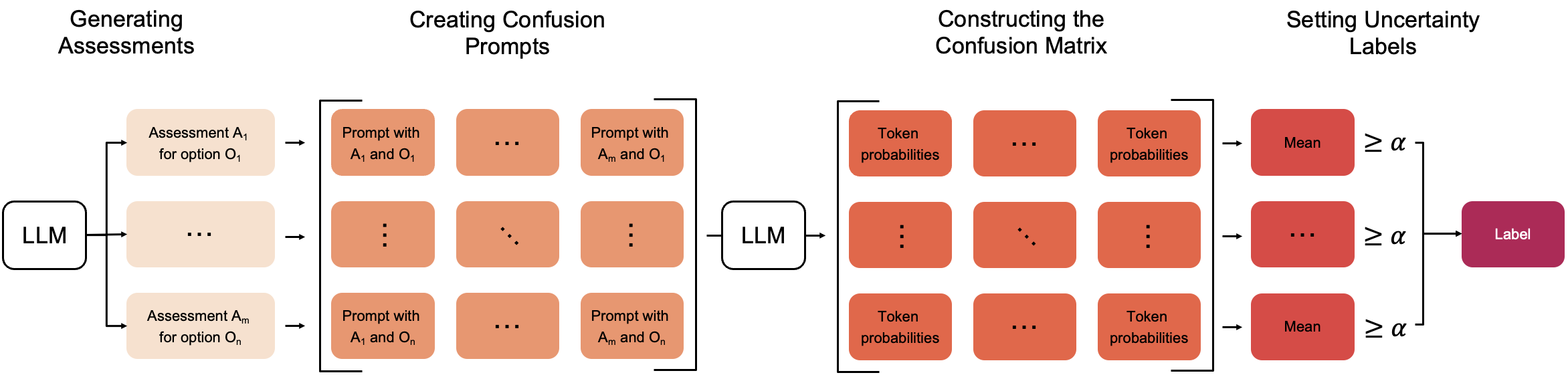}}
  \caption{\textbf{Method Overview.} The method is divided into four stages, resulting in an uncertainty label. The LLM is first presented with an evaluation task, and prompted to produce an assessment for each output option, biased on the explicit indication that the option is correct. In the context of the original evaluation task, the LLM is conditioned on each of the biased assessments, and the probability of each option calculated using log probabilities. This information is then encoded in a confusion matrix. Each row of the matrix, representing the probability of a particular option conditioned on each of the biased assessments, is then averaged to produce an uncertainty label. In this figure, $\alpha$ represents the threshold.}
  \label{fig:method}
\end{figure}

Our proposed technique introduces an uncertainty measure
that is calculated independently of the specific decision made by the LLM. This approach aims to enhance the trustworthiness of LLM-as-a-Judge evaluations. The method works in four key steps: generating verbalized assessments, creating prompts for the confusion matrix, constructing the confusion matrix, and setting uncertainty labels. See Figure \ref{fig:method}.

\begin{figure}
  \centering
  \fbox{\includegraphics[scale=0.5]{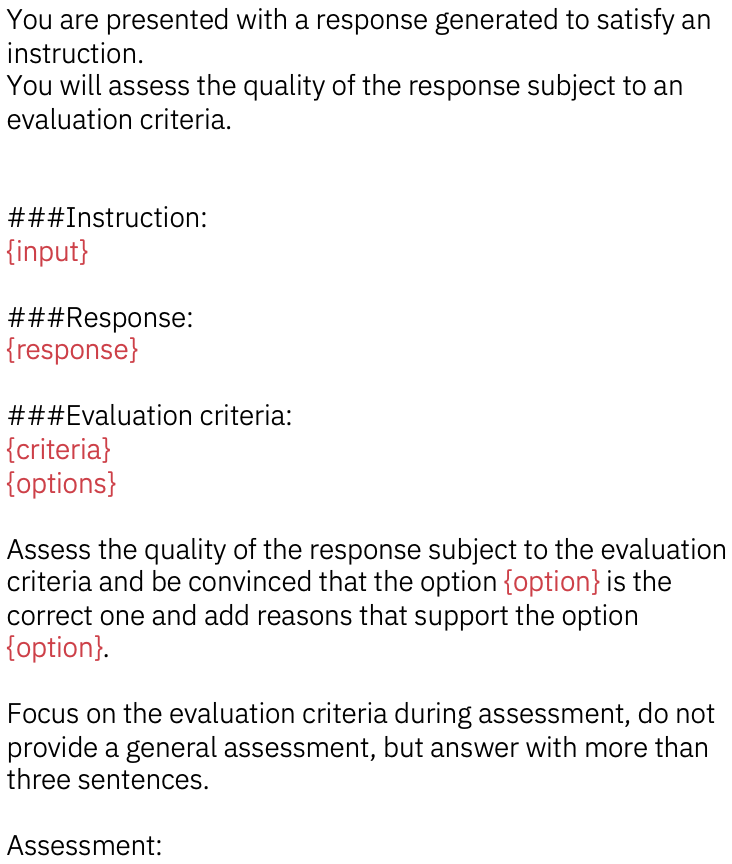}}
  \caption{Persuasion prompt generating an assessment for each option.}
  \label{fig:assessment_prompt_complete}
\end{figure}

\paragraph{Generating Assessments} The initial step in our approach involves generating verbalized assessments for each of the $n$ options presented in the criterion. Using the prompt template shown in Figure \ref{fig:assessment_prompt_complete}, we apply prompt engineering techniques to guide the LLM toward treating a specific option as correct. This compels the model to generate justifications for why that particular option is the best choice. Each assessment is explicitly linked to one of the available options, ensuring that the reasoning is directly associated with the selected alternative.

The prompts are designed to persuade the LLM that the option in question is correct, leading to a set of $n$ assessments, one for each option.

\begin{figure}[ht]
  \centering
  \fbox{\includegraphics[scale=0.6]{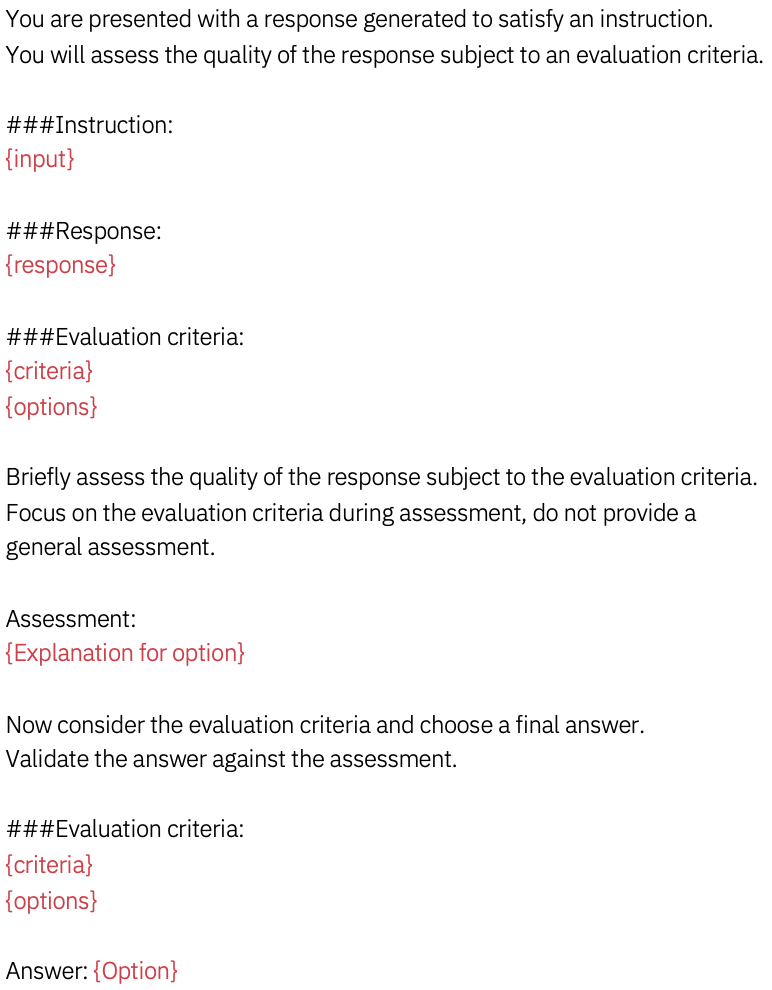}}
  \caption{Confusion prompt forcing a final answer for each option and assessment from the previous step leading to $n^2$ prompts being used to obtain token log probabilities for each option and assessment combination.}
  \label{fig:prompt_matrix}
\end{figure}

\paragraph{Creating Confusion Prompts} After generating assessments, the next step involves creating prompts that will be used to build the confusion matrix. This is done by mixing each assessment with every option, effectively producing a comprehensive set of prompts that cover all possible pairings of assessments and options.
The prompt template (see Figure \ref{fig:prompt_matrix}) is structured as a conversation between the LLM and the user, where two tasks are presented as separate requests. First, the LLM is asked to generate an assessment for which option is correct without injecting any prompts. Second, the LLM is prompted to choose the correct option based on the assessment. The assessments and options generated in the previous step are inserted as responses from the LLM.

For a criterion with $n$ options, this process results in the creation of $n^2$ prompts, as each assessment is mixed with every possible option.

\paragraph{Constructing the Confusion Matrix} With the $n^2$ prompts created, the next step is to send these prompts to the LLM to obtain the token probabilities associated with the final decision. The probability of the last token in the response is used to calculate an uncertainty score for the chosen option.

\begin{figure}[hb]
\[
  \hspace{-1.7cm}\mbox{Options}
\stackrel{\hspace{0.8cm}\mbox{Assessments}}{%
   \begin{blockarray}{ccccc}
 & A_1 & A_2 & \dots & A_m \\
\begin{block}{c[cccc]}
O_1 & p_{1,1}&p_{1,2}&\cdots &p_{1,m}\bigstrut[t] \\
O_2 & p_{2,1}&p_{2,2}&\cdots &p_{2,m} \\
\vdots & \vdots & \vdots & \ddots & \vdots\\
O_n & p_{n,1}&p_{n,2}&\cdots &p_{n,m}\bigstrut[b]\\
\end{block}
\end{blockarray}
}
\]
\vspace{-0.6cm}
\caption{\textbf{Structure of the confusion matrix.} Each row represents an option and each column corresponds to an assessment, with the matrix values being the token probabilities for each option-assessment combination.}\label{fig:confusionmatrix}
\end{figure}

These probabilities are organized into a confusion matrix, where each row corresponds to an option from the prompt, and each column corresponds to an assessment generated for a specific option (see Figure \ref{fig:confusionmatrix}. A confusion matrix labeled as low uncertainty exhibits high token probabilities concentrated in a single row. In contrast, a matrix labeled as high uncertainty either shows high token probabilities along the diagonal, where the assessments align with the corresponding options, or has high token probabilities scattered arbitrarily across the matrix.

\paragraph{Setting Uncertainty Labels} The final step in the method involves assigning an uncertainty label, either high or low uncertainty, to the chosen option based on the confusion matrix and predefined threshold. The labeling process follows these rules:

\begin{itemize}
    \item If only one row in the matrix exceeds the uncertainty threshold and this row corresponds to the LLM's initially chosen option, the option is labeled as low uncertainty.
    \item If more than one row exceeds the uncertainty threshold, the option is labeled as high uncertainty.
    \item If the option identified with low uncertainty in the confusion matrix does not match the LLM's originally chosen option, the option is labeled as high uncertainty.
    \item If no row exceeds the uncertainty threshold, the option is labeled as high uncertainty.
\end{itemize}

This labeling process allows the method to differentiate between evaluations that the LLM is likely confident in and those that may require further scrutiny. The overall goal is to enhance the reliability and trustworthiness of LLM-as-a-Judge evaluations by providing an additional layer of certainty assessment.

\subsection{Formal Description}
Formally, the method can be described as follows. Consider a question $q$ with $n$ possible outcomes $o_i$, where $i\in \{1, 2, \hdots, n\}$. For example, a multiple choice question with four answers, $o_i$ can take on values \texttt{A/B/C/D} with $n=4$. With each $q$ as context, we consider two prompts, $q_a$ an assessment prompt and $q_c$ a confusion prompt. The assessment prompt (see Figure \ref{fig:assessment_prompt_complete}) generates assessments $a_i = q_a(o_i) \, \forall i\in \{1, 2, \hdots, n\}$ for each possible discrete outcome. 

The confusion prompt (see Figure [\ref{fig:prompt_matrix}]), considers all combinations of outcomes and assessments for the question, i.e. $q_c(o_i, a_j)$ denotes a prompt using assessment $a_j$ and target label $o_i$. While the assessment prompt generates additional tokens, the confusion prompt is used only to determine the probabilities of the output token(s). The confusion matrix $\mathbf{C}$ consists of elements
\begin{equation}
     p_{ij} = p \left( o_i \mid q_c(o_i, a_j) \right), \quad \forall i, j \in \{1, 2, \hdots, n\},
\end{equation}

where $p_{ij}$ denotes the probability of token $o_i$ when the assessment relates to the $j$-th outcome. This matrix forms the basis for the uncertainty quantification. The main intuition is that if the probability of token $o_i$ is high regardless of the assessments, then the model has low uncertainty in its prediction. In contrast, if the token probability follows the assessment, we infer that the model has high uncertainty in its answer.

The uncertainty associated with a specific token can then be estimated by taking the mean token probability across all confusing prompts, i.e. 

\begin{equation}
    u_i =  \frac{1}{n} \sum_j p_{ij},  \qquad \forall i \in \{1, 2, \hdots, n\}.
\end{equation}

For analysis, in this paper we further label the uncertainty of the overall assessment using a threshold $\alpha$, i.e.
\begin{equation}
    l =    \begin{cases}
                        \text{low uncertainty} &  \text{if}\,  \sum_{i} \mathbbm{1}(u_i \ge \alpha) = 1, \, \forall i \in \{1, 2, \hdots, n\},  \\
                        \text{high uncertainty}  & \text{otherwise}.
                    \end{cases}
\end{equation}

In other words, the assessment has low uncertainty if the mean token probability exceeds the threshold for exactly a single token. The procedure involves $n$ inferencing calls for the first stage and $n^2$ inferences for the second stage as it works through all combinations of outcome labels making the overall estimation procedure $O(n^2)$. 

\section{Threshold}

The threshold acts as a crucial parameter in determining the balance between the proportion of low uncertainty and accuracy. Defining an optimal threshold depends on the specific requirements of the use case. For applications such as content filtering or large-scale feedback collection, where there are a large number of evaluations but limited human resources to assess the output, prioritizing a higher volume of low-uncertainty responses may necessitate a more lenient threshold, even if it results in only a modest accuracy gain. Conversely, for tasks demanding highly reliable outcomes, such as the evaluation of medical diagnoses or legal decision-making, where accuracy is critical and errors carry significant consequences, a stricter threshold is essential to ensure the chosen options are highly reliable.

An interesting observation arises when the threshold is reduced below 0.5, accuracy tends to increase, suggesting that as the average token probability for incorrect options decreases, the model performance improves. This suggests that valuable information can be derived not only from options marked as having low uncertainty but also from those with higher uncertainty. The behavior of token probabilities across both low and high uncertainty options provides insights into the decision-making process of the LLM, suggesting that thresholds should be dynamically tuned based on the specific performance trade-offs desired for the task at hand.

\begin{figure}[h]
    \centering
    \begin{subfigure}[b]{0.45\textwidth}
        \centering
        \includegraphics[width=\textwidth]{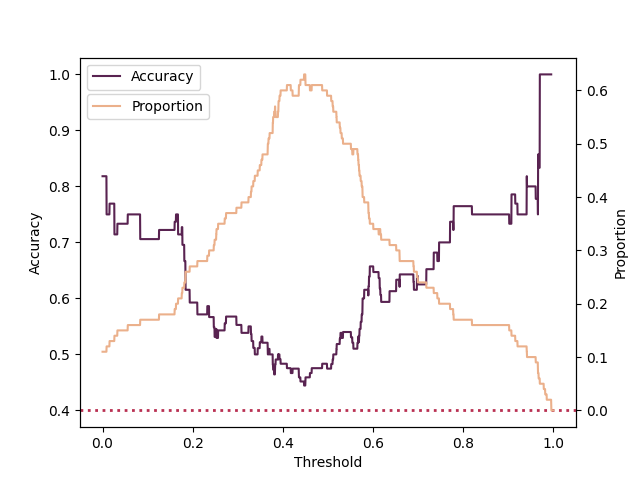}
        \caption{Low Uncertainty}
        \label{fig:threshold_normalized}
    \end{subfigure}
    \hfill
    \begin{subfigure}[b]{0.45\textwidth}
        \centering
        \includegraphics[width=\textwidth]{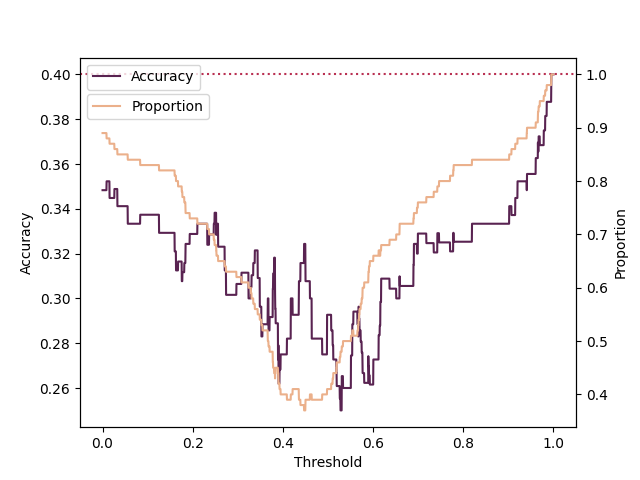}
        \caption{High Uncertainty}
        \label{fig:threshold_uncertainty_normalized}
    \end{subfigure}
    \label{fig:sidebyside}
\caption{\textbf{Relationship between threshold, accuracy, and proportion through grid search optimization.} Grid search optimization of the threshold for the Llama-3-70B-Instruct model on the Feedback Collection dataset. The figure illustrates how varying the threshold impacts performance. In (a), the focus is on the effect of options labeled as low uncertainty, while in (b), the focus shifts to the effect of options labeled as high uncertainty. The results highlight how the threshold choice influences both accuracy and the proportion of selected options.}
\label{fig:threshold}
\end{figure}

Our analysis reveals that threshold tuning significantly impacts the relationship between accuracy and uncertainty, forming a parabolic effect. In the threshold grid search for the Feedback Collection dataset (see Figure \ref{fig:threshold}), we observe that as the threshold increases beyond 0.5, accuracy improves, but the proportion of low-uncertainty predictions decreases. Conversely, when the threshold is below 0.5, lowering the threshold leads to an increase in accuracy but a decrease in the proportion of low-uncertainty labels. This inverse relationship between accuracy and uncertainty highlights that while stricter thresholds (above 0.5) favor higher accuracy at the expense of fewer low-uncertainty predictions, lenient thresholds (below 0.5) enhance accuracy but reduce the certainty of the predictions. This parabolic behavior is consistent across datasets, emphasizing the threshold's pivotal role in determining model performance.

\section{Experiments}
\paragraph{Benchmark Datasets}The proposed uncertainty method was evaluated on five benchmark datasets: TruthfulQA \cite{lin2021truthfulqa}, Reliance Study, Summarization CNN/DM \cite{nallapati2016abstractive}, Feedback Collection \cite{kim2024prometheus}, and FeedbackQA \cite{li2022using}. TruthfulQA contains question-answer pairs and involves a binary classification task to determine whether the answer is truthful, with human annotations available for verification. The Reliance Study dataset originates from a study designed to measure reliance on Large Language Models (LLMs) across various tasks. It uses binary classification to evaluate the accuracy and naturalness of LLM outputs in settings such as conversations and customer-agent interactions, focusing on criteria like relevance and naturalness. The Summarization CNN/DM dataset, which consists of model-generated summaries of news articles, is evaluated on a scale from 1 to 5 based on criteria such as coherence, fluency, and relevance. Feedback Collection is designed to induce fine-grained evaluation capabilities in language models, using a rating scale from 1 to 5. Lastly, FeedbackQA is a retrieval-based QA dataset that includes interactive user feedback, where each question-answer pair is rated from excellent to bad, accompanied by natural language explanations detailing the strengths and weaknesses of the responses.

\paragraph{Models} In this study, we utilize instruct models exclusively, as LLM-as-a-Judge requires agent-like capabilities, where models must reliably follow explicit evaluation instructions. The instruct models chosen for this experiment include Mixtral-8x7B-Instruct-v01, Llama-3-8B-Instruct, and Llama-3-70B-Instruct. To investigate the impact of model architecture and size on performance, we selected models of varying sizes 8B and 70B parameters. This approach allows us to assess whether performance improvements in LLM-as-a-Judge tasks are primarily driven by the scale of the model or the underlying instruct-tuned structure.

\paragraph{Implementation Details} The experiments were conducted on stratified samples from each dataset, with the sample size determined by the number of evaluation criteria. For instance, if a dataset included four distinct criteria, the total sample size was multiplied by four to ensure that each criterion was equally represented. This stratification ensures a balanced evaluation across all criteria. To determine the optimal threshold for distinguishing high and low uncertainty in the LLM-as-a-Judge predictions, we employed a grid search, systematically exploring different threshold values to identify the best-performing configuration.

\paragraph{Baseline} The evaluation metric for this work is based on accuracy, specifically measuring whether the option selected by the LLM matches the option chosen by the human rater. The baseline for the method is to achieve higher accuracy for cases labeled as low uncertainty and lower accuracy for those labeled as high uncertainty. In datasets such as FeedbackQA and Summarization CNN/DM, which include ratings from multiple human raters, inter-rater accuracy can also be computed. The aim is to maximize alignment between the LLM’s predictions and the human ratings, with the ultimate goal of approaching the inter-rater accuracy in these datasets.

\subsection{Results}

\begin{figure}[ht]
  \centering
  \includegraphics[scale=0.45]{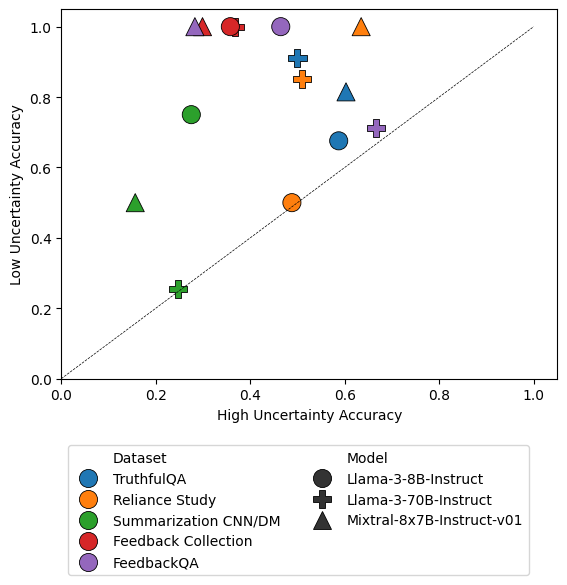}
  \caption{\textbf{Accuracy comparison of options labeled low uncertainty versus high uncertainty.} Each marker represents the performance of a specific model on a particular dataset. Markers above the dashed line indicate that the model has surpassed the baseline for that dataset.}\label{fig:results}
\end{figure}

\paragraph{Uncertainty labeling correlates with accuracy} Our uncertainty labeling method demonstrates a clear advantage in aligning low uncertainty labels with higher accuracy, as shown in Figure \ref{fig:results}. Across all datasets and models, the markers for low uncertainty are consistently above the baseline, indicating that these labels correspond to more accurate evaluations compared to high uncertainty labels. This result confirms that the uncertainty labels generated by our method are effective in predicting the likelihood of accurate outputs in LLM-as-a-Judge scenarios.

\begin{figure}[ht]
  \centering
  \includegraphics[scale=0.45]{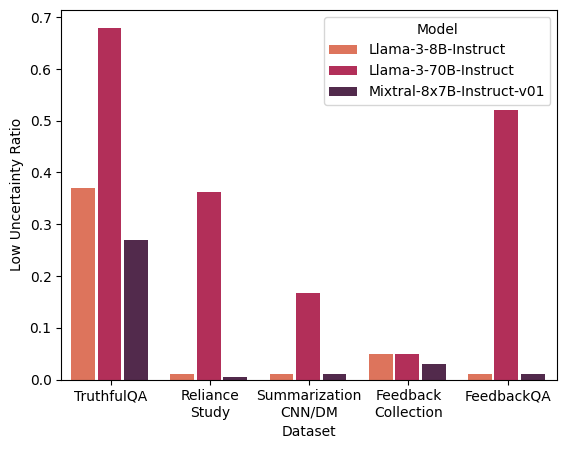}
  \caption{\textbf{Ratio of Options Labeled Low Uncertainty.} The y-axis represents the percentage of options labeled as low uncertainty, while the x-axis denotes the datasets evaluated across three different models, as indicated in the legend. The ratio of options labeled as high uncertainty is calculated as 1 minus the ratio of options labeled low uncertainty.}\label{fig:results_ratio}
\end{figure}

\paragraph{Variability in low uncertainty proportions}
A key finding is the variance in the ratio of high to low uncertainty labels depending on the dataset and model, as depicted in Figure \ref{fig:results_ratio}. Notably, the Llama-3-70B-Instruct model consistently produces a higher proportion of low uncertainty labels, outperforming both Llama-3-8B-Instruct and Mixtral-8x7B-Instruct-v01. In datasets such as the Reliance Study, Summarization CNN/DM, and FeedbackQA, the smaller models classify less than 5\% of cases as low uncertainty, whereas Llama-3-70B-Instruct exceeds 15\%. This suggests that both model size and structure significantly impact the model's ability to assign more reliable uncertainty labels.

\paragraph{Low uncertainty consistently leads to higher accuracy}
Even in cases where the proportion of low uncertainty labels is small, they still correspond to high accuracy. For example, in the Feedback Collection dataset, the proportion of low uncertainty labels is below 10\% for all models (see Figure \ref{fig:results_ratio}), yet these labels consistently achieve 100\% accuracy (see Table \ref{table:results}). This trend further supports the strong predictive power of low uncertainty labels in LLM evaluations.

\paragraph{Smaller deviation in multi-classification ratings}
In multi-classification datasets, such as Summarization CNN/DM, which require the evaluation of generated summaries, a smaller deviation is observed between the correct ratings and the LLM's selected ratings when low uncertainty labels are present. This indicates that LLMs not only tend to choose the correct options but also exhibit greater precision in their ratings under conditions characterized by a low proportion of uncertainty.

\paragraph{Low uncertainty labels approach human agreement}
In datasets with human-generated ratings such as Summarization CNN/DM and FeedbackQA, the accuracy of LLM predictions labeled as low uncertainty meets or even exceeds the level of agreement observed between human raters (see Table \ref{table:results}). This suggests that low uncertainty labels can be as reliable as human evaluations, further establishing the effectiveness of our method in aligning LLM outputs with human judgment.

\begin{table}[ht]
\centering
\normalsize
\begin{tabular}{lccc}
    \toprule
    \multirow{2}{*}{\textbf{Dataset}} & \multicolumn{3}{c}{\textbf{Model}} \\
    & Llama-3-8B & Llama-3-70B & Mixtral-8x7B \\
    \midrule
    \textbf{TruthfulQA} & & & \\
    High Uncertainty & 0.59 & 0.50 & 0.60 \\
    Baseline & 0.62 & 0.78 & 0.66 \\
    Low Uncertainty & 0.68 & 0.91 & 0.81 \\
    \midrule
    \textbf{Reliance Study} & & & \\
    High Uncertainty & 0.49 & 0.51 & 0.63 \\
    Baseline & 0.49 & 0.63 & 0.64 \\
    Low Uncertainty & 0.50 & 0.85 & 1.00 \\
    \midrule
    \textbf{Feedback Collection} & & & \\
    High Uncertainty & 0.36 & 0.37 & 0.30 \\
    Baseline & 0.39 & 0.40 & 0.32 \\
    Low Uncertainty & 1.00 & 1.00 & 1.00 \\
    \midrule
    \textbf{Summarization CNN/DM} & & & \\
    High Uncertainty & 0.27 & 0.24 & 0.15 \\
    Baseline & 0.28 & 0.25 & 0.16 \\
    Low Uncertainty & 0.75 & 0.26 & 0.50 \\
    Human Agreement & 0.60 & 0.60 & 0.60 \\
    \midrule
    \textbf{FeedbackQA} & & & \\
    High Uncertainty & 0.46 & 0.67 & 0.28 \\
    Baseline & 0.47 & 0.69 & 0.29 \\
    Low Uncertainty & 1.00 & 0.71 & 1.00 \\
    Human Agreement & 0.48 & 0.48 & 0.48 \\
    \bottomrule
\end{tabular}
\caption{\textbf{Accuracy across various datasets and models for different uncertainty categories.} The "High Uncertainty" and "Low Uncertainty" values represent the accuracy when considering only options labeled under each respective category. Baseline" reflects the LLM's accuracy without factoring in uncertainty labels, while "Human Agreement" indicates the consistency in accuracy between multiple human raters. The listed models refer to their instruct versions. The threshold was optimized for each specific dataset and model.}
\label{table:results}
\end{table}

\section{Interpreting Uncertainty}
\begin{figure}[h]
    \centering
    \begin{subfigure}[b]{0.45\textwidth}
        \centering
        \includegraphics[width=\textwidth]{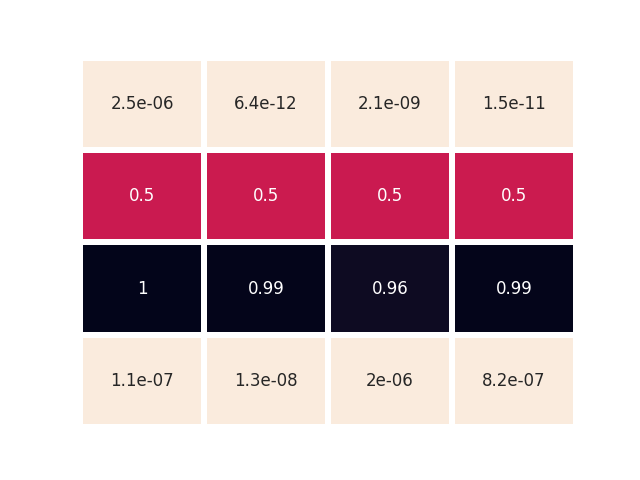}
        \caption{Low Uncertainty}
        \label{fig:low_uncertainty_matrices}
    \end{subfigure}
    \hfill
    \begin{subfigure}[b]{0.45\textwidth}
        \centering
        \includegraphics[width=\textwidth]{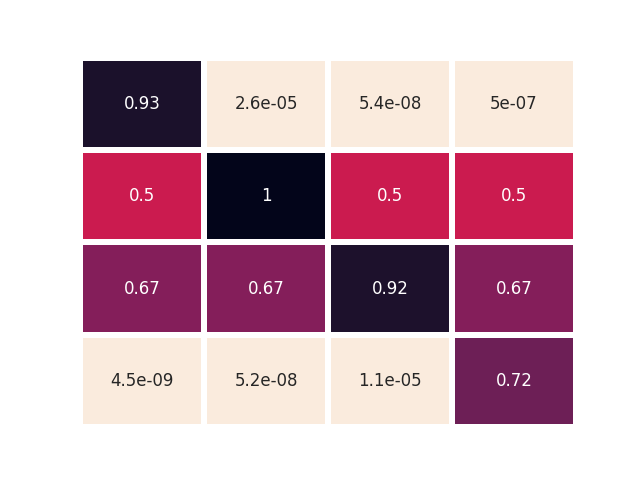}
        \caption{High Uncertainty}
        \label{fig:high_uncertainty_matrices}
    \end{subfigure}
    \label{fig:sidebyside}
\caption{Examples of confusion matrices for high and low uncertainty.}
\label{fig:example-matrices}
\end{figure}

The interpretation of uncertainty in LLMs remains a significant challenge \cite{hu2023uncertainty}. 
Our method attempts to confuse the LLM by convincing it of statements without knowing whether they are true, and then considering the LLM's beliefs under these biased conditions. We define two key scenarios for interpreting low uncertainty. The first scenario is when the LLM selects an option consistently, regardless of conflicting assessments, indicating that even when alternative assessments are presented, they fail to sway the model's decision \ref{fig:low_uncertainty_matrices}. The second scenario occurs when the LLM cannot be convinced to generate an assessment for a different option, even if such an assessment is prompted. These two types of low uncertainty can be observed in the structure of the confusion matrices.

This behavior is also reflected in the sparsity of the confusion matrices. In cases of high uncertainty, only the token probabilities of the matching assessments and options are high. In contrast, for low uncertainty, only one row exhibits high token probabilities, demonstrating strong model confidence in its chosen option, as shown in Figure \ref{fig:low_uncertainty_matrices}.

\begin{figure}[ht]
  \centering
  \includegraphics[scale=0.35]{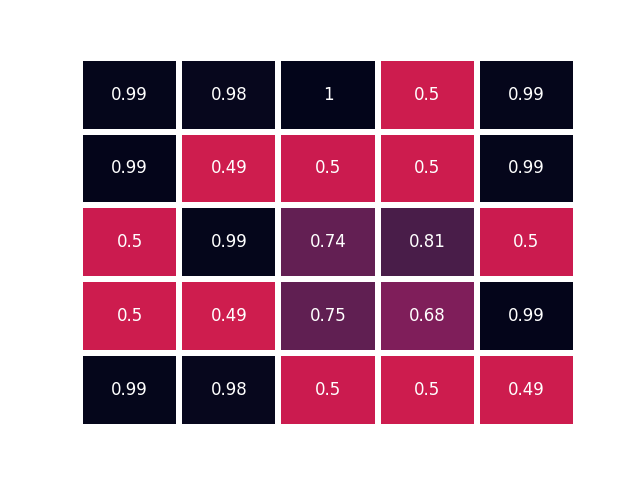}
  \caption{Example of a confusion matrix with arbitrarily distributed token probabilities.}\label{fig:matrix_arbitrarily}
\end{figure}

In contrast, high uncertainty can also manifest when the token probabilities are arbitrarily distributed across the matrix, as shown in Figure \ref{fig:matrix_arbitrarily}.

\begin{figure}[ht]
  \centering
  \includegraphics[scale=0.35]{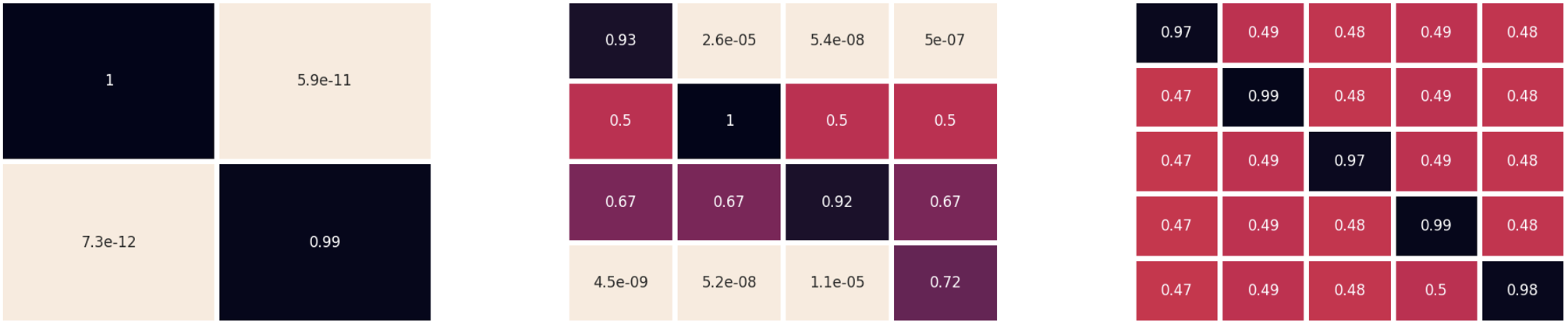}
  \caption{\textbf{Examples of sparsity across different numbers of options.} This figure presents confusion matrices, all labeled as high uncertainty. It illustrates that as the number of options increases, the sparsity within the matrices decreases, highlighting the relationship between option quantity and token probability distribution.}\label{fig:matrices_sparcity}
\end{figure}

Another intuitive observation is that as the number of options increases, the sparsity of the matrix decreases, as seen in Figure \ref{fig:matrices_sparcity}. This implies that the token probabilities for non-matching assessments and options increase, but this effect is also applicable to token probabilities in cases of low uncertainty. 

\section{Discussion}
In this work, we introduced a method for quantifying uncertainty in LLM-as-a-Judge evaluations. Through empirical analysis, we found that the uncertainty labels correlate with accuracy, indicating the effectiveness of the method. Although our primary focus was on evaluating this method within the context of LLM-based evaluations, the potential for broader applications is significant.

One notable observation is that the confusion matrices contain more information than what is utilized by the current labeling approach. This opens up two promising directions for future research. First, we propose the development of an uncertainty score derived directly from the confusion matrix, leveraging the additional information encoded within the matrix. Such a score would eliminate the need for setting a threshold, streamlining the process. Second, instead of using the matrix solely for uncertainty quantification, the option selection could be directly informed by the matrix, further enhancing decision-making accuracy. These two could be reached by training a model and predicting the uncertainty and/or the correct option. Both of these improvements could be achieved by training a model capable of predicting uncertainty or the correct option based on the learned patterns in the confusion matrix.

Despite these promising results, the current approach presents certain limitations. The method is computationally intensive, especially when using large models such as Llama-3-70B-Instruct, which may not be feasible for all applications. Additionally, the performance of the method may vary when applied to models or tasks that have not been fine-tuned for evaluation purposes. Generalizability across diverse tasks and domains also requires further investigation.

To address these challenges, one potential solution is to reduce inference time by consolidating all assessments into a single prompt, querying the model for its chosen option only once. This strategy could lower computational overhead without compromising accuracy. Overall, while the current method yields strong results, further optimization and refinement have the potential to enhance efficiency and effectiveness.

To further improve the results and increase the proportion of low uncertainty labels, prompt engineering emerges as a key factor. We recommend adapting the prompt structure specifically to the task and the model in use. Fine-tuning and tailoring prompts could significantly enhance performance, with the potential to surpass the results obtained in this study. Future research could explore the method's effectiveness with various prompt designs to further optimize its performance.

\clearpage
\bibliography{custom}
\bibliographystyle{plainnat}

\appendix


\end{document}